%% file: main.tex
\documentclass{article} 
\ifdefined\localfontfix \pdfmapline{+phvb TeXGyreHeros-Bold "encqec ReEncodeFont" <q-ec.enc <qhvb.pfb}\fi
\usepackage{iclr2027_conference}
\IfFileExists{ptmr8t.tfm}{\usepackage{times}}{\usepackage{newtxtext}}

\input{math_commands.tex}

\usepackage[utf8]{inputenc}
\usepackage[T1]{fontenc}
\usepackage{hyperref}
\usepackage{url}
\usepackage{amsmath}
\usepackage{amssymb}
\usepackage{mathtools}
\usepackage{amsthm}
\usepackage{graphicx}
\usepackage{booktabs}
\usepackage{array}
\usepackage{float}
\usepackage{multirow}
\usepackage{xcolor}
\usepackage{tikz}

\definecolor{exampleback}{HTML}{F1F2F1}
\definecolor{findingback}{HTML}{FFF5D6}
\definecolor{findingframe}{HTML}{D9A92B}
\definecolor{principleback}{HTML}{F2ECFA}
\definecolor{principleframe}{HTML}{8662C4}
\newsavebox{\jevboxcontent}
\newcommand{\jevboxskip}{\medskip}
\newenvironment{jevbox}[3]{\par\jevboxskip\noindent\def\jevboxframe{#1}\def\jevboxback{#2}\setlength{\fboxsep}{6pt}\setlength{\fboxrule}{0.6pt}%
  \begin{lrbox}{\jevboxcontent}\begin{minipage}{\dimexpr\linewidth-2\fboxsep-2\fboxrule\relax}\textbf{#3.}\ \ignorespaces}%
  {\end{minipage}\end{lrbox}\fcolorbox{\jevboxframe}{\jevboxback}{\usebox{\jevboxcontent}}\par\jevboxskip}
\newcounter{example}
\newenvironment{examplebox}[1]{\refstepcounter{example}\begin{jevbox}{exampleback}{exampleback}{Example~\theexample\ (#1)}\small}{\end{jevbox}}
\newenvironment{findingbox}[1]{\renewcommand{\jevboxskip}{\smallskip}\begin{jevbox}{findingframe}{findingback}{#1}}{\end{jevbox}}

\definecolor{promptback}{HTML}{EEF3FA}
\definecolor{promptframe}{HTML}{7D9BC1}

\newenvironment{promptspec}[1]{\begin{jevbox}{promptframe}{promptback}{#1}\small\par\smallskip\noindent\begin{tabular}{@{}>{\bfseries\raggedright\arraybackslash}p{2.35cm}@{\hspace{6pt}}>{\raggedright\arraybackslash}p{\dimexpr\linewidth-2.35cm-6pt\relax}@{}}}{\end{tabular}\end{jevbox}}
\usetikzlibrary{arrows.meta,positioning,fit,calc}
\usepackage[font=small,labelfont=bf]{caption}
\usepackage{subcaption}

\theoremstyle{definition}
\newtheorem{definition}{Definition}

\newcommand{\jev}{Jev}

\definecolor{vizblue}{HTML}{2A78D6}
\definecolor{vizaqua}{HTML}{1BAF7A}
\definecolor{vizorange}{HTML}{EB6834}
\definecolor{vizgray}{HTML}{8A8984}

\title{Code Owns the Simulation, Jev Owns the Evaluation}

\author{\setlength{\tabcolsep}{0pt}%
\begin{tabular}[t]{@{}l@{}}
{\small Yaodong Yang$^{1}$\hspace{0.5em}Hongyao Tang$^{2}$\hspace{0.5em}Yi Ma$^{3}$\hspace{0.5em}Xingyu Fan$^{1}$\hspace{0.5em}Weixun Wang$^{4}$\hspace{0.5em}Jinpeng Li$^{5}$\hspace{0.5em}Tianpei Yang$^{6}$} \\[6pt]
{\normalfont\small $^{1}$The Chinese University of Hong Kong\enspace $^{2}$Tianjin University\enspace $^{3}$Shanxi University\enspace $^{4}$Independent Researcher} \\
{\normalfont\small $^{5}$CAIR, Hong Kong Institute of Science \& Innovation, Chinese Academy of Sciences\enspace $^{6}$Nanjing University} \\[3pt]
{\normalfont\small \texttt{yydapple@gmail.com}, \texttt{tanghongyao@tju.edu.cn}, \texttt{mayi@sxu.edu.cn}, \texttt{xyfan@link.cuhk.edu.hk},} \\
{\normalfont\small \texttt{wxwang@tju.edu.cn}, \texttt{jinpeng.li@cair-cas.org.hk}, \texttt{tianpei.yang@nju.edu.cn}}
\end{tabular}}

\iclrfinalcopy
\begin{document}

\maketitle
\lhead{Preprint}

\input{sections/0_abstract}
\input{sections/1_intro}
\input{sections/3_framework}

\input{sections/4_crt}
\input{sections/5_games}
\input{sections/6_text}
\input{sections/7_dmc}
\input{sections/2_related}
\input{sections/8_discussion}

\section*{Reproducibility Statement}
Appendix~\ref{app:interface} gives the instruction of every condition and the model version, and Appendix~\ref{app:probes} gives the probe questions. The other appendices describe the CRT items, the game generator, the behavioral models, the ALFWorld labeler and lookahead, the DMC skills and the knowledge probes. Every model call is stored in a raw run file. One script recomputes every number, figure and table of the CRT, game and ALFWorld studies from these files. The DMC numbers come from the run files of the DMC evaluation scripts, and the frames of Figure~\ref{fig:dmc} come from a script that replays the logged decisions in the simulator. We will release the code, the raw runs and these scripts. \jev{} is accessed through a public API, so its outputs can change across versions. Repeating a single call changes its probabilities by 0.04 on average.

\bibliography{references}
\bibliographystyle{iclr2027_conference}

\clearpage
\appendix
\input{sections/appendix}

\end{document}

%% file: math_commands.tex
\usepackage{amsmath,amsfonts,bm}

\def\eqref#1{equation~\ref{#1}}

\def\1{\bm{1}}

\DeclareMathAlphabet{\mathsfit}{\encodingdefault}{\sfdefault}{m}{sl}
\SetMathAlphabet{\mathsfit}{bold}{\encodingdefault}{\sfdefault}{bx}{n}



%% file: sections/0_abstract.tex
\begin{abstract}
Judgment models such as \jev{} return, in a single call and without reasoning text, a probability for each described option. This makes them attractive as an agent's action-selection layer, but it is unclear which decisions they can be trusted with. We test \jev{} on reflection tests, one-shot matrix games, the text game ALFWorld and robot control, and find a sharp boundary. \jev{} succeeds when the right option can be judged from the information in the input, which we call \emph{evaluation}. Specifically, it solves 99\% of the counterintuitive Cognitive Reflection Test questions. However, it fails when the right option depends on \emph{simulation} (i.e., predicting something not in the input), such as the opponent's action or the subgoal that must come first. In games, \jev{} plays suboptimally as if its rational opponent acted at random when the opponent's action is not given. In ALFWorld, \jev{} favors commands whose object or place is mentioned in the task description. For example, given the task ``put a clean knife in the drawer'', \jev{} carries an unwashed knife straight to the drawer instead of first washing it at the sink. Surprisingly, many of these failures are not due to a lack of knowledge. Asked separately what the opponent will do, \jev{} usually answers correctly, and it responds well given the opponent's action. It fails when one call must both perform the simulation and evaluate based on it. This suggests letting code make the prediction or simulation. When code supplies it, such as a lookahead in ALFWorld and physics simulation in robot control, \jev{} becomes an expert controller through its general evaluation ability.
\end{abstract}

%% file: sections/1_intro.tex
\section{Introduction}
\label{sec:intro}

Many decisions of an agent are choices within a short list of options, such as the next command in a text game, the next move in a game or the next skill of a robot. Judgment models such as \jev{}, built as a general ``System One'', target exactly this kind of choice \citep{typesafe2026jev}. In a single call and without reasoning text, \jev{} returns a probability for each described option. It needs no task-specific training and no output parsing. This makes \jev{} attractive as an agent's action-selection layer, called at every step as a fast policy over listed actions.

\input{sections/fig_framework}

But it is unclear \emph{which} kind of decisions such a model can be trusted with. Benchmark scores do not provide the answer. For example, \jev{} alone solves 31\% of ALFWorld games, but this number says how often it fails, not at which decisions. Two common views of such models do not provide the answer either. The first view comes from the ``System One'' label in the dual-process theory of \citet{kahneman2011thinking} and \citet{evans2013dual}, which suggests failure wherever an intuitive option is wrong. Yet \jev{} solves 99\% of the counterintuitive Cognitive Reflection Test questions, which are built around an intuitive but wrong answer \citep{frederick2005cognitive,hagendorff2023human}. The second view holds that small models lack reasoning. Yet \jev{} plays the best response in 98\% of matrix games once the opponent's action is given.

We test \jev{} on reflection tests, one-shot matrix games, the text game ALFWorld \citep{shridhar2021alfworld} and robot control, and find a sharp boundary. Specifically, we find that \jev{} succeeds when the right option can be judged from the information in the input, which we call \emph{evaluation}. For example, in a CRT question, a bat and a ball cost \$1.10 in total, and the bat costs \$1.00 more. By checking the listed prices against the question, \jev{} selects \$0.05 for the ball and avoids the intuitive \$0.10.

However, \jev{} fails when the right option depends on \emph{simulation} (i.e., predicting something not in the input), such as the opponent's action or the subgoal that must come first. In RL terms, \jev{} behaves like a policy without a world model. It scores actions by the current input and does not look ahead. In games, \jev{} plays suboptimally as if its rational opponent acted at random when the opponent's action is not given. In ALFWorld, we observe that \jev{} favors commands whose object or place is mentioned in the task description. Its accuracy is 99\% when such a command is correct, but 36\% when such commands are wrong. For example, given the task ``put a clean butterknife in drawer'', \jev{} carries an unwashed butterknife straight to the drawer instead of first washing it at the sink (Figure~\ref{fig:hero}c).

Surprisingly, we find that many of these failures are not due to a lack of knowledge. Asked separately what the opponent will do, \jev{} usually answers correctly (90\%), and it plays the best response 98\% of the time given the opponent's action. We therefore attribute the failure to a single call in which \jev{} must both perform the simulation and evaluate based on it.

This leads us to let code make the prediction or simulation. When code supplies it, \jev{} becomes an expert controller through its general evaluation ability. In one-shot games, giving \jev{} the opponent's action raises accuracy from 33\% to 98\% when the action with the highest average payoff is not the best response to a rational opponent. Even when the opponent's action is not supplied, \jev{} can predict it in one call. Given this prediction in a second call, \jev{} plays the best action in 94\% of cases. Similarly, with a lookahead simulation in ALFWorld, solved games rise from 31\% to 87\% (Table~\ref{tab:alfworld} in Appendix~\ref{app:alfworld}). Finally, in robot control, \jev{} never picks up a box when it chooses raw torques. In contrast, with skills for which code simulates the physical dynamics, it matches a hand-written controller on the native task and further generalizes to new task variants.

Our contributions are as follows.
\begin{itemize}\setlength{\itemsep}{1pt}\setlength{\topsep}{2pt}
\item \textbf{Diagnosis.} Across reflection tests, one-shot games, ALFWorld and robot control, we find a sharp boundary. \jev{} succeeds at evaluation but follows surface shortcuts when the right option depends on simulation (Sections~\ref{sec:crt}--\ref{sec:dmc}).
\item \textbf{Cause.} We show that many of these failures are not due to a lack of knowledge. \jev{} fails when one call must both perform the simulation and evaluate based on it. Separating simulation and evaluation recovers from these failures (Sections~\ref{sec:games-probe} and~\ref{sec:text-model}).
\item \textbf{Remedy.} We propose that code owns the simulation and \jev{} owns the evaluation. With this division, \jev{} becomes an expert controller in games, ALFWorld and robot control (Sections~\ref{sec:games-probe}, \ref{sec:text-lookahead} and~\ref{sec:dmc}).
\end{itemize}

%% file: sections/fig_framework.tex
\begin{figure}[t]
\centering
\begin{tikzpicture}[font=\fontsize{6.2}{7.4}\selectfont, x=1cm, y=1cm,
  card/.style={draw=vizgray!60, rounded corners=3pt, fill=white, align=center, inner sep=2pt},
  arr/.style={-{Stealth[length=4pt]}, vizgray!90!black, line width=0.6pt},
  line/.style={vizgray!90!black, line width=0.6pt},
  txt/.style={anchor=west, align=left, inner sep=1pt, execute at begin node={\hyphenpenalty=10000\exhyphenpenalty=10000}},
  head/.style={anchor=west, inner sep=1pt, font=\fontsize{6.2}{7.4}\selectfont\itshape, text=vizgray!80!black},
  ico/.style={line width=0.75pt, line cap=round, line join=round},
  dk/.style={draw=vizgray!55!black}]
\newcommand{\icstate}[1][0.62]{\tikz[ico, scale=#1]{\draw[dk, fill=white, rounded corners=1pt] (0,0) -- (0,1) -- (0.58,1) -- (0.82,0.76) -- (0.82,0) -- cycle;
  \draw[dk, fill=vizblue!15] (0.58,1) -- (0.58,0.76) -- (0.82,0.76) -- cycle;
  \foreach \y/\w in {0.62/0.38, 0.42/0.26, 0.22/0.32} {\fill[vizblue, rounded corners=0.6pt] (0.12,\y-0.055) rectangle (0.3,\y+0.055); \draw[vizgray!55!black] (0.38,\y) -- (0.38+\w,\y);}}}
\newcommand{\icoptions}[1][0.62]{\tikz[ico, scale=#1]{\foreach \y/\f in {0.83/vizblue, 0.5/white, 0.17/white} {\draw[draw=vizblue, fill=vizblue!10, rounded corners=2pt] (0,\y-0.13) rectangle (1,\y+0.13);
  \draw[draw=vizblue, fill=\f] (0.15,\y) circle (0.065); \draw[vizgray!55!black] (0.32,\y) -- (0.82,\y);}}}
\newcommand{\iccode}[1][0.62]{\tikz[ico, scale=#1]{\draw[dk, fill=white, rounded corners=1.5pt] (0,0) rectangle (1.1,0.9); \fill[vizgray!55!black, rounded corners=1.5pt] (0,0.68) rectangle (1.1,0.9);
  \foreach \x in {0.1,0.22,0.34} \fill[white] (\x,0.79) circle (0.035);
  \draw[draw=vizorange] (0.38,0.52) -- (0.2,0.34) -- (0.38,0.16); \draw[draw=vizorange] (0.72,0.52) -- (0.9,0.34) -- (0.72,0.16); \draw[draw=vizorange] (0.6,0.54) -- (0.5,0.14);}}
\newcommand{\icoutcome}[1][0.62]{\tikz[ico, scale=#1]{\fill[vizgray!55!black] (0,0.5) circle (0.08); \draw[dk] (0,0.5) -- (0.3,0.5);
  \draw[dk, rounded corners=2pt] (0.3,0.5) |- (0.62,0.85); \draw[dk] (0.3,0.5) -- (0.62,0.5); \draw[dk, rounded corners=2pt] (0.3,0.5) |- (0.62,0.15);
  \foreach \y/\c/\d in {0.85/vizaqua/vizaqua!70!black, 0.5/vizgray!45/vizgray, 0.15/vizorange/vizorange!70!black} \draw[draw=\d, fill=\c, rounded corners=1pt] (0.62,\y-0.12) rectangle (1.02,\y+0.12);}}
\newcommand{\icinput}[1][0.62]{\tikz[ico, scale=#1]{\draw[draw=vizblue, fill=vizblue!12, rounded corners=1.5pt] (0,0) -- (0,0.9) -- (0.4,0.9) -- (0.5,0.78) -- (1.1,0.78) -- (1.1,0) -- cycle;
  \draw[draw=vizblue] (0.2,0.52) -- (0.6,0.52); \draw[draw=vizblue] (0.2,0.3) -- (0.75,0.3);}}
\newcommand{\iclure}[1][0.62]{\tikz[ico, scale=#1]{\draw[dk, fill=vizorange] (0.05,0.95) -- (0.05,0.47) arc[start angle=180, end angle=360, radius=0.45] -- (0.95,0.95) -- (0.69,0.95) -- (0.69,0.47) arc[start angle=0, end angle=-180, radius=0.19] -- (0.31,0.95) -- cycle;
  \draw[dk, fill=vizgray!25] (0.05,0.95) rectangle (0.31,0.76); \draw[dk, fill=vizgray!25] (0.69,0.95) rectangle (0.95,0.76);}}
\newcommand{\iccheck}[1][0.62]{\tikz[ico, scale=#1]{\fill[vizaqua] (0.5,0.5) circle (0.5); \draw[white, line width=1.3pt] (0.26,0.5) -- (0.43,0.32) -- (0.75,0.68);}}
\newcommand{\icdrawer}[1][0.5]{\tikz[ico, scale=#1]{\draw[dk, fill=vizorange!12, rounded corners=1pt] (0,0) rectangle (1,1); \draw[dk] (0,0.5) -- (1,0.5);
  \draw[draw=vizorange, line width=1.1pt] (0.38,0.75) -- (0.62,0.75); \draw[draw=vizorange, line width=1.1pt] (0.38,0.25) -- (0.62,0.25);}}
\newcommand{\icsink}[1][0.5]{\tikz[ico, scale=#1]{\draw[dk, fill=vizblue!15] (0.08,0.45) arc[start angle=180, end angle=360, radius=0.42] -- cycle; \draw[dk] (0,0.45) -- (1,0.45);
  \draw[dk, rounded corners=1.5pt] (0.5,0.45) -- (0.5,0.95) -- (0.78,0.95) -- (0.78,0.82); \fill[vizblue] (0.78,0.72) circle (0.06); \fill[vizblue] (0.78,0.58) circle (0.045);}}
\newcommand{\acc}[4]{{\renewcommand{\arraystretch}{1.05}\setlength{\tabcolsep}{1.6pt}\def\l##1{{\fontsize{4.6}{5.2}\selectfont\color{vizgray!80!black}##1}}\def\n##1{{\fontsize{5.8}{6.6}\selectfont\bfseries\color{#1}##1}}%
  \color{vizgray!60}\setlength{\arrayrulewidth}{0.4pt}\begin{tabular}{|c|c|c|}\hline \l{games} & \l{ALFWorld} & \l{robot}\\\hline \n{#2} & \n{#3} & \n{#4}\\\hline\end{tabular}}}
\node[anchor=west, font=\footnotesize\bfseries] at (0,0.3) {(a) \jev{} alone};
\fill[vizgray!8, rounded corners=3pt] (0,-0.05) rectangle (4.0,-4.0);
\node[card, minimum width=1.7cm, minimum height=1.25cm, text width=1.56cm] (A1) at (1.0,-0.825) {\icstate\\State $x$\\{\tiny task, history}};
\node[card, minimum width=1.7cm, minimum height=1.25cm, text width=1.56cm] (A2) at (3.0,-0.825) {\icoptions\\Options\\{\tiny candidate actions}};
\node[card, minimum width=3.7cm, minimum height=0.75cm, align=none, draw=vizorange, fill=vizorange!8] (A3) at (2.0,-2.125)
  {$\vcenter{\hbox{\includegraphics[height=0.600cm]{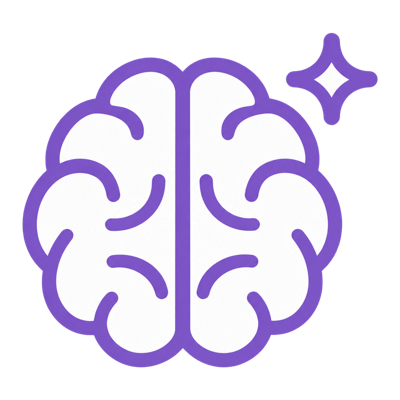}}}$\hspace{4pt}\begin{tabular}{@{}l@{}}\textbf{\jev{}, one call}\\must simulate \emph{and} evaluate\end{tabular}};
\node[card, minimum width=3.7cm, minimum height=1.05cm, align=none] (A4) at (2.0,-3.325)
  {$\vcenter{\hbox{\iclure[0.55]}}$\hspace{5pt}\begin{tabular}{@{}c@{}}\textbf{Picks the surface shortcut}\\[2pt]\acc{vizorange!85!black}{33\%}{36\%}{0/20}\end{tabular}};
\draw[arr] (A1.south) -- (A1.south |- A3.north);
\draw[arr] (A2.south) -- (A2.south |- A3.north);
\draw[arr] (A3) -- (A4);
\node[anchor=west, font=\footnotesize\bfseries] at (4.3,0.3) {(b) Code owns the simulation, \jev{} owns the evaluation (ours)};
\fill[exampleback, rounded corners=3pt] (4.3,-0.05) rectangle (13.85,-1.8);
\fill[principleback, rounded corners=3pt] (4.3,-2.1) rectangle (13.85,-4.0);
\node[card, minimum width=1.3cm, minimum height=1.4cm, text width=1.16cm] (S) at (5.1,-0.9) {\icstate\\State $x$};
\node[card, minimum width=1.3cm, minimum height=1.4cm, text width=1.16cm] (O) at (6.75,-0.9) {\icoptions\\Options};
\node[card, minimum width=2.25cm, minimum height=1.4cm, text width=2.11cm, draw=promptframe, fill=promptback] (C) at (8.875,-0.9)
  {\iccode\\Simulator in code\strut\\[1pt]{\tiny\itshape\color{promptframe!70!black}\renewcommand{\arraystretch}{0.78}\begin{tabular}{@{}c@{}}game rules, a copy\\of the game, physics\end{tabular}}};
\node[card, minimum width=3.35cm, minimum height=1.4cm, text width=3.21cm, draw=promptframe, line width=0.7pt] (R) at (12.025,-0.9)
  {\icoutcome\\Outcome of each option\strut\\[1pt]{\tiny\itshape\color{promptframe!70!black}\renewcommand{\arraystretch}{0.78}\begin{tabular}{@{}c@{}}opponent's action, next\\observation, skill result\end{tabular}}};
\draw[arr] (S) -- (O);
\draw[arr] (O) -- (C);
\draw[arr] (C) -- (R);
\node[card, minimum width=2.95cm, minimum height=1.6cm, text width=2.81cm] (I) at (5.925,-3.05) {\icinput\\Input\\{\tiny state, options, outcomes}};
\node[card, minimum width=3.6cm, minimum height=1.6cm, inner xsep=3pt, draw=principleframe, line width=0.8pt] (J) at (9.55,-3.05)
  {$\vcenter{\hbox{\includegraphics[height=0.792cm]{figures/icons/jev.png}}}$\hspace{3pt}\begin{tabular}{@{}l@{}}\textbf{\jev{}, one call} \\[0.5pt] {\tiny\itshape\color{principleframe!80!black}judges with world knowledge} \\[3pt] Which option is best,\\given its outcome? $\to p(a)$\end{tabular}};
\node[card, minimum width=2.0cm, minimum height=1.6cm, text width=1.86cm] (A) at (12.7,-3.05)
  {\iccheck[0.5]\\[1pt]\textbf{Best option}\\[2pt]\acc{vizaqua!70!black}{98\%}{79\%}{9/20}};
\draw[line, rounded corners=2pt] (S.south) |- (R.south |- 0,-1.95) -- (R.south);
\draw[line] (O.south) -- (O.south |- 0,-1.95);
\draw[arr] (I.north |- 0,-1.95) -- (I.north);
\draw[arr] (I) -- (J);
\draw[arr] (J) -- (A);
\node[anchor=west, font=\footnotesize\bfseries] at (0,-4.35) {(c) One ALFWorld decision: {\mdseries ``put a clean butterknife in drawer'', holding an} unwashed {\mdseries knife}};
\fill[vizgray!8, rounded corners=3pt] (0,-4.6) rectangle (13.85,-6.95);
\fill[principleback, rounded corners=2pt] (10.45,-4.67) rectangle (12.35,-6.88);
\foreach \x/\t in {0.15/Option (2 of 42), 3.45/Outcome simulated by code (abbreviated), 8.75/\jev{} alone}
  \node[head] at (\x,-4.87) {\t};
\node[head, text=principleframe!90!black, font=\fontsize{6.2}{7.4}\selectfont\bfseries\itshape] at (10.5,-4.87) {\jev{} with code};
\draw[vizgray!40] (0.15,-5.05) -- (13.7,-5.05);
\draw[vizgray!40] (0.15,-5.95) -- (13.7,-5.95);
\foreach \y/\icon/\opt/\why/\out/\pa/\pb/\col/\role in {
  -5.5/\icdrawer/{go to drawer 1}/{names the destination}/{``The drawer 1 is closed.'' Then: open drawer 1.}/0.60/0.07/vizorange/lure,
  -6.42/\icsink/{go to sinkbasin 1}/{wash the knife first}/{``You see a lettuce 3, \ldots'' Then: clean butterknife 2 with sinkbasin 1.}/0.17/0.83/vizaqua/correct} {
  \node[anchor=west, inner sep=0pt] at (0.15,\y) {\icon};
  \node[txt, text width=2.4cm] at (0.85,\y) {\texttt{\opt}\\{\itshape\color{\col!80!black}\why}};
  \node[txt, text width=5.0cm] at (3.45,\y) {\out};
  \fill[vizblue] (8.75,\y-0.07) rectangle ++(1.0*\pa,0.14);
  \node[anchor=west, inner sep=1pt] at (8.75+1.0*\pa+0.03,\y) {\pa};
  \fill[vizaqua] (10.5,\y-0.07) rectangle ++(1.2*\pb,0.14);
  \node[anchor=west, inner sep=1pt] at (10.5+1.2*\pb+0.03,\y) {\pb};
  \node[rounded corners=2pt, fill=\col!18, text=\col!70!black, font=\fontsize{6.2}{7.4}\selectfont\bfseries, minimum width=1.1cm] at (13.1,\y) {\role};
}
\end{tikzpicture}
\caption{\textbf{Code owns the simulation, and \jev{} owns the evaluation.} (a) Alone, one \jev{} call must both predict what each option leads to and evaluate the options. \jev{} then picks the surface shortcut. Accuracy is 33\% in misleading one-shot games and 36\% at misleading ALFWorld steps, and with raw torques \jev{} solves none of 20 robot episodes. (b) In the grey layer, code simulates each option with the game rules, a copy of the game or a physics engine, and writes the outcome next to the option. In the purple layer, \jev{} reads the state, the options and their outcomes, and evaluates the options in one call. Accuracy rises to 98\% in the games and 79\% at the ALFWorld steps, and \jev{} solves 9 of the 20 robot episodes. (c) One ALFWorld decision. Alone, \jev{} goes to the drawer, which the task mentions. With the outcome of each command, it goes to the sink to wash the knife first.}
\label{fig:hero}
\end{figure}

%% file: sections/3_framework.tex
\section{Evaluation, Simulation, and How We Label Decisions}
\label{sec:framework}

This section defines evaluation and simulation and explains how we label each decision.

\textbf{Setting.} \jev{} (version \texttt{jev-1.13.0}) takes a JSON state and a list of candidate options, each with a text description \citep{typesafe2026jev}. It returns the chosen option, a probability for every option and a confidence, but no reasoning text. We use it zero-shot, with one fixed instruction per experimental condition (Appendix~\ref{app:interface}). A decision is a triple $(s, \mathcal{A}, a^\star)$, where $s$ is the state, $\mathcal{A}$ the set of described candidate options and $a^\star \subseteq \mathcal{A}$ the correct ones.

\begin{definition}[Evaluation and simulation]
A decision needs only \emph{evaluation} if $a^\star$ can be judged from the information that $s$ and the candidate options provide. On the other hand, a decision depends on \emph{simulation} if finding $a^\star$ requires a prediction $z$ with information not in the input. The prediction can be shared by all candidate options, like the opponent's action or the next subgoal. It can also belong to one option, like the next observation after executing that option. If code supplies $z$ by adding it to $s$, the decision needs only evaluation. Code can obtain $z$ by running a simulator, such as the game rules, a copy of the environment or a physics engine, or by asking the model in a separate call.
\end{definition}

Whether a decision needs simulation depends on how the task is presented, not on the model (Example~\ref{ex:sink}).

\begin{examplebox}{one decision, presented two ways}\label{ex:sink}
When the agent holds an unwashed butterknife, going to the sink is right because the knife can be cleaned there. If an option only names a command, such as ``go to sinkbasin 1'', the fact that the knife can be cleaned there must be simulated by code or predicted by the model (Figure~\ref{fig:hero}c, \jev{} alone). If an option also shows what the command makes possible, such as ``Then: clean butterknife 2 with sinkbasin 1'', the same decision needs only evaluation (Figure~\ref{fig:hero}c, \jev{} with code).
\end{examplebox}

Knowing that a decision depends on simulation, however, does not separate real success from a lucky guess. A model can get such a decision right when a surface shortcut happens to point to the correct candidate option. Consider the task ``put a clean butterknife in drawer''. Once the butterknife is clean, going to the drawer is correct, and a model that simply follows the words of the task gets it right. We therefore define such a surface shortcut for each domain.

\begin{definition}[Surface shortcut and lure]
For each domain, the \emph{surface shortcut} $h$ is a simple rule, defined in advance, that picks the candidate options that look most relevant from the input alone, without predicting anything. A decision that depends on simulation is \emph{aligned} if $h$ picks a correct candidate option and \emph{misleading} otherwise. In a misleading decision, the candidate options that $h$ picks are the \emph{lure}.
\end{definition}

In games, $h$ picks the action with the highest average payoff, as if the opponent acted at random. This is the level-1 rule of behavioral game theory \citep{stahl1995players,costagomes2001cognition}. In ALFWorld, $h$ picks the commands that mention an object or place named in the task description. Following \citet{hagendorff2023human}, we classify each answer to a misleading decision as \emph{correct}, \emph{lure} or \emph{other}. If most errors are caused by lures, the model is being drawn to the surface shortcut. If errors are spread over the other wrong options, they are more likely random noise.

\textbf{Predictions.} Our claim concerns what \jev{} does, not how it works internally. If \jev{} evaluates but does not simulate, five things should hold. (i) Decisions that need only evaluation are answered well, even against an intuitive but wrong option. (ii) Aligned decisions that depend on simulation are answered well, because the surface shortcut picks a correct candidate option. (iii) Misleading decisions that depend on simulation fail, and the errors go to the lure. (iv) Stating the missing prediction $z$ in the input, such as the opponent's action, restores accuracy. (v) If \jev{} can predict $z$ in a separate call, writing its answer back also restores accuracy. \textbf{Prediction (v)} tells whether \jev{} cannot predict $z$ at all, or can predict it correctly when asked separately but fails when it must also make the choice in the same call. By contrast, if intuitive lures were enough to mislead \jev{}, it would fail the CRT, whose questions are built around intuitive but wrong answers, against \textbf{prediction (i)}. If \jev{} simply lacked reasoning, giving it the opponent's action would not help, against \textbf{prediction (iv)}.

\textbf{Protocol.} The order of the candidate options shifts \jev{}'s probabilities, so we average each decision over several orders. Whether a decision needs only evaluation or depends on simulation, and whether it is aligned or misleading, are computed from the environment, never from \jev{}'s choices.

%% file: sections/4_crt.tex
\section{Jev Owns the Evaluation: Cognitive Reflection}
\label{sec:crt}

\begin{table}[H]
\centering
\small
\caption{\textbf{\jev{} avoids the intuitive answer on the CRT.} Fractions over the 150 questions, each in four option orders. In the control versions, the intuitive answer is correct. Published results are open-ended answers, shown for reference only. The 95\% interval of \jev{}'s correct rate on lure questions is [0.97, 1.00].}
\label{tab:crt}
\setlength{\tabcolsep}{4pt}
\input{tables/crt.tex}
\end{table}

\textbf{Setup.} We first test \jev{} on decisions that need only evaluation. Does an intuitive but wrong option make \jev{} fail? The Cognitive Reflection Test (CRT) puts \textbf{prediction (i)} to a strong test, because its questions are built around intuitive but wrong answers, so that the first answer that comes to mind is wrong. We use the 150 new questions of \citet{hagendorff2023human}. Each question becomes four candidate options, namely the correct answer, the intuitive answer and two other wrong answers, shown in four orders.

\textbf{Results.} \jev{} picks the correct answer on 99\% of questions and the intuitive answer on 1\% (Table~\ref{tab:crt}). For reference, people answering the same questions in open form give the intuitive answer 55\% of the time. Each question also has a control version without the trap, in which the intuitive answer is correct. \jev{} is right on 96\% of these, so it does not simply avoid intuitive-looking options. Because the answers are listed as options, \jev{} only needs to check each one against the conditions in the question (Example~\ref{ex:bat}). Such checking needs only evaluation, so \textbf{prediction (i)} holds: an intuitive lure alone does not make \jev{} fail. Next, we further validate this boundary on one-shot matrix games, where the right action can depend on a prediction of the opponent's action.

\begin{examplebox}{bat and ball}\label{ex:bat}
A bat and a ball cost \$1.10 in total, and the bat costs \$1.00 more than the ball. The intuitive answer for the ball is \$0.10. Checking the listed options against the question gives the answer. A ball of \$0.05 gives a bat of \$1.05 and a total of \$1.10, as the question states, so \jev{} picks \$0.05.
\end{examplebox}

%% file: tables/crt.tex
\begin{tabular*}{\linewidth}{@{\extracolsep{\fill}}lccccc}
\toprule
 & Four options & \multicolumn{4}{c}{Published, open-ended answers \citep{hagendorff2023human}} \\
\cmidrule(lr){2-2}\cmidrule(lr){3-6}
 & \jev{} & Humans (n=455) & GPT-3 davinci-003 & ChatGPT-3.5 & ChatGPT-4 \\
\midrule
Lure questions, correct & \textbf{0.99} & 0.38 & 0.05 & 0.59 & 0.96 \\
Lure questions, intuitive & 0.01 & 0.55 & 0.80 & 0.15 & 0.00 \\
\midrule
Control versions, correct & 0.96 & -- & -- & -- & -- \\
\bottomrule
\end{tabular*}

%% file: sections/5_games.tex
\section{Jev Does Not Own the Simulation: One-Shot Games}
\label{sec:games}

We now test \jev{} on decisions that depend on simulation. Can \jev{} predict what the opponent will do, and how does \jev{} act on it? One-shot games let us control exactly what must be simulated. The payoff table is fully visible, but the value of each action depends on the opponent's action, which the table does not describe. We vary how many steps of reasoning about the opponent the answer needs and whether the surface shortcut happens to pick the correct action.

\subsection{Failures follow the surface shortcut, not depth}
\label{sec:games-acc}

\textbf{Setup.} We sample random $3{\times}3$ matrix games with distinct payoffs in 0--99 that iterated elimination of strictly dominated strategies solves to a unique outcome. \jev{} plays Row against a Column described in the state as rational and informed. The \emph{depth} $D$ is the number of elimination rounds Row needs, minus one. $D{=}0$ means Row has a dominant action, which pays more than every other action whatever Column plays. At $D{=}k$, Row must reason $k$ steps about Column, such as ``Column will never play this action, so I should play that one''. A game is \emph{aligned} if Row's best action against a rational Column, the equilibrium action, is also the action with the highest average payoff, and \emph{misleading} otherwise. We sample 30 games for $D{=}0$ and for each combination of $D{=}1$, $2$, $3$ with aligned or misleading, resulting in 210 games in total. Each game is shown with Row's three actions in three different orders. We compare three conditions. In the \emph{plain} condition, \jev{} sees only the payoff table. In the \emph{reason} condition, the instruction also tells \jev{} to first rule out the actions that Column would never play. In the \emph{told} condition, code adds Column's equilibrium action to the state, as in ``Column will play Y.'', which tests \textbf{prediction (iv)}.

\begin{figure}[t]
\centering
\begin{minipage}[t]{0.42\linewidth}
\vspace{-1.1mm}
\begin{tikzpicture}[font=\scriptsize, x=1cm, y=1cm]
\node[anchor=west, font=\footnotesize\bfseries] at (0,0.88) {(a) A misleading game, $D{=}1$};
\begin{scope}[xshift=0.25cm]
\node at (1.62,0.47) {Column};
\node[rotate=90] at (-0.1,-0.725) {Row};
\foreach \c/\x in {X/0.78, Y/1.62, Z/2.46} \node at (\x,0.15) {\c};
\fill[vizgray!20] (2.04,-0.05) rectangle (2.88,-1.4);
\foreach \r/\y/\a/\b/\cc in {A/-0.275/{24, 5}/{71, 83}/{12, 94}, B/-0.725/{54, 21}/{2, 9}/{83, 34}, C/-1.175/{86, 56}/{85, 0}/{62, 99}} {
  \node at (0.2,\y) {\r};
  \node at (0.78,\y) {\a}; \node at (1.62,\y) {\b}; \node at (2.46,\y) {\cc};
}
\draw[vizgray!70] (0.36,-0.05) rectangle (2.88,-1.4);
\foreach \x in {1.2,2.04} \draw[vizgray!70] (\x,-0.05) -- (\x,-1.4);
\foreach \y in {-0.5,-0.95} \draw[vizgray!70] (0.36,\y) -- (2.88,\y);
\draw[vizaqua!70!black, thick] (0.36,-0.5) rectangle (2.88,-0.95);
\draw[vizorange, thick] (0.36,-0.95) rectangle (2.88,-1.4);
\node at (3.3,0.15) {Avg.}; \node at (3.95,0.15) {Plain}; \node at (4.6,0.15) {Told};
\foreach \y/\m/\p/\t in {-0.275/35.7/0.08/0.00, -0.725/46.3/0.13/0.96, -1.175/\textbf{77.7}/0.79/0.04} {
  \node at (3.3,\y) {\m}; \node at (3.95,\y) {\p}; \node at (4.6,\y) {\t};
}
\node[anchor=west, text=vizaqua!65!black] at (4.9,-0.725) {eq.};
\node[anchor=west, text=vizorange] at (4.9,-1.175) {lure};
\end{scope}
\node[anchor=north west, text width=5.7cm, align=left, inner sep=0pt, font=\scriptsize, execute at begin node={\hyphenpenalty=10000\exhyphenpenalty=10000}] at (0,-1.84) {Cells give Row's payoff, then Column's. Z pays Column the most in every row, so a rational Column plays Z (shaded). Against Z, B pays Row the most. The surface shortcut picks C, which has the highest average payoff. Plain and Told give \jev{}'s probabilities, averaged over three option orders.};
\end{tikzpicture}
\end{minipage}\hfill
\begin{minipage}[t]{0.56\linewidth}
\vspace{0pt}
\centering
{\footnotesize\bfseries (b) Accuracy over all 210 games}\\[2pt]
\includegraphics[width=\linewidth]{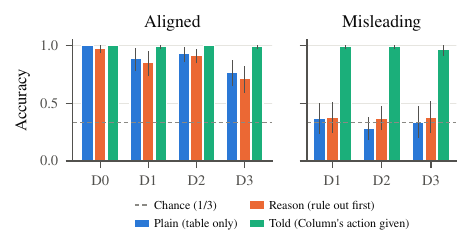}
\end{minipage}
\caption{\textbf{One-shot games: failures follow the surface shortcut, not depth.} (a) An example game. (b) Accuracy is how often \jev{} chooses the equilibrium action (the right answer). Error bars are 95\% bootstrap intervals over games. In aligned games the surface shortcut picks the equilibrium, and in misleading games it does not. Plain, reason and told are the three conditions of Section~\ref{sec:games-acc}. Reason does not help, while told restores accuracy at every depth.}
\label{fig:games}
\end{figure}

\textbf{Results.} Figure~\ref{fig:games}b shows that alignment, not depth, determines accuracy. With a dominant action ($D{=}0$), which is best whatever Column plays and so needs only evaluation, \jev{} is right in 100\% of games. In aligned games it stays at 77--93\% for every depth. In misleading games it falls to chance (28--37\%) and picks the lure in 53--71\% of games. A regression confirms that depth has no detectable effect once alignment is taken into account (Appendix~\ref{app:pilot}). An instruction to reason leaves misleading games at chance (37--38\%). When code supplies Column's action in the state, accuracy in these games rises to 97--99\%. This confirms \textbf{predictions (i)--(iv)}.

\begin{findingbox}{Contribution 1: diagnosis}
\jev{} succeeds at decisions that need only evaluation. When the right option depends on simulation, it follows the surface shortcut, and its errors go to the lure.
\end{findingbox}

\subsection{What Jev computes}
\label{sec:games-model}

Which decision rule produces these choices? We fit \jev{}'s probabilities with simple rules that score every action, and a softmax with one free parameter turns the scores into probabilities (Appendix~\ref{app:pilot}). The rule that scores each action by its average payoff fits best. This rule is a soft version of the surface shortcut. It picks the action with the highest average payoff with a probability that grows with its margin over the next-best action. Adding whether an action is the equilibrium barely improves the fit. \jev{} thus plays as if its rational opponent acted at random, a quantal level-1 player in the terms of behavioral game theory \citep{mckelvey1995quantal,stahl1995players}. Level 1 involves no reasoning about the opponent, so it is not the same as depth $D{=}1$.

\subsection{Knowing is not using}
\label{sec:games-probe}

Does \jev{} fail because it cannot predict Column's action, or because it cannot use that prediction while choosing? Following \citet{hagendorff2023human}, we ask for the prediction in a separate call on the same games. \jev{} correctly predicts Column's equilibrium action 90\% of the time, mostly by picking the column that pays Column the most on average (216 of its 244 correct predictions). Yet \jev{}'s own choice does not use this prediction. We split the plain decisions by whether \jev{}'s separate prediction for the opponent's action in the same game was right. \jev{} chooses correctly in 32\% of the decisions where its prediction was right and in 35\% of those where it was wrong. Including a right prediction in the state makes the choice right in almost all calls, as in the told condition (98\%). \jev{} thus fails when one call must both perform the simulation and evaluate based on it.

Separating simulation and evaluation recovers the decision. Code first asks \jev{} to predict Column's action and then writes the answer into the state, as in the told condition. Accuracy in misleading games rises from 33\% to 94\%, as \textbf{prediction (v)} expects. The first call predicts Column's action from Column's average payoff. The second call only evaluates Row's actions against the written prediction. This chains two level-1 steps into level-2 play, the best response to a level-1 opponent. The choice follows the written prediction. When \jev{}'s own prediction is wrong, code writes a wrong action into the state, and \jev{} instead plays the best response to that wrong prediction in 24 of 26 calls.

\begin{findingbox}{Contribution 2: cause}
\jev{} can often make the prediction and make correct decisions based on the prediction, but it fails when one call must do both.
\end{findingbox}

The one-shot matrix games isolate the failure, but they are small and artificial. In contrast, a household agent faces long episodes and dozens of commands, and the missing prediction may be a step that the task never states, such as washing a butterknife before putting it away. The next section asks whether the same failure appears in more realistic settings.

%% file: sections/6_text.tex
\section{Code Owns the Lookahead: Text Games}
\label{sec:text}

ALFWorld \citep{shridhar2021alfworld} tests the same boundary in a sequential task. Its household tasks, such as ``put a clean butterknife in drawer'', are text games in which each step is a choice among admissible commands. Playing on its own, \jev{} solves only 31\% of the 134 unseen games. We ask where it fails, what it knows and whether letting code perform the simulation fixes it.

\subsection{Where Jev fails}
\label{sec:text-diag}

\textbf{Setup.} To test single decisions, we replay the successful trajectories from ALFWorld's hand-coded expert. At each step \jev{} sees the task, the observation, the full history and the admissible commands. A choice is correct if it lies in an \emph{acceptable set} derived from the environment. For example, when the expert goes to sinkbasin 1 to wash the knife, going to sinkbasin 2 also counts as correct. We label each step with one of four types that we define in advance and compute automatically from the task and the history (Appendix~\ref{app:alfworld}). \textbf{E} steps directly complete part of the task, such as taking the butterknife, cleaning it or putting it in the drawer. Whether such a command is right can be read from the command itself, so these steps need only evaluation. \textbf{S-known} steps complete no part of the task themselves but make a later step possible, such as going to the sink while holding an unwashed butterknife. They depend on simulation, although all information needed for them is in the history. \textbf{S-search} steps explore for an object not yet seen, so the needed information is missing from both the history and the current state. Therefore, they also depend on simulation. However, the object may be in any receptacle not yet checked, so many commands are acceptable there and even random choice scores high (48--74\%, Table~\ref{tab:alfworld-full} in Appendix~\ref{app:alfworld}). We therefore report these steps only in the appendix. \textbf{Other} steps, mostly looking around or closing a receptacle, neither complete part of the task nor make a later step possible, so no command is clearly right or wrong there. We exclude them (22\% of expert steps). The main text thus covers E and S-known steps. We test on both of ALFWorld's evaluation splits, called seen and unseen. The 140 seen games take place in rooms that also appear in ALFWorld's training data, and the 134 unseen games take place in new rooms. We use \jev{} zero-shot on both. The surface shortcut splits S steps into aligned and misleading (Section~\ref{sec:framework}). We defined the ALFWorld surface shortcut according to the unseen games, such as picking ``go to drawer 1'' for the task ``put a clean butterknife in drawer'', and then tested it on the seen games (Appendix~\ref{app:alfworld}).

\begin{figure}[t]
\centering
\begin{minipage}[t]{0.63\linewidth}
\vspace{0pt}
{\footnotesize\bfseries (a) One trajectory: ``put a clean butterknife in drawer''}\\[3pt]
\scriptsize
\setlength{\tabcolsep}{2.5pt}
\begin{tabular}{@{}clll@{}}
\toprule
Step & Step type & Expert command & \jev{}'s top choice \\
\midrule
1 & E & take butterknife 2 & same (0.98) \\
2 & \textcolor{vizorange}{S-known, misleading} & go to sinkbasin 1 & \textcolor{vizorange}{go to drawer 1 (0.60)} \\
3 & E & clean butterknife 2 with sinkbasin 1 & same (0.96) \\
4 & \textcolor{vizaqua!65!black}{S-known, aligned} & go to drawer 1 & same (0.69) \\
5 & E & move butterknife 2 to drawer 1 & same (1.00) \\
\bottomrule
\end{tabular}\\[3pt]
{\scriptsize Consecutive decisions of the expert's trajectory (opening the drawer omitted). \texttt{go to drawer 1} is the lure before the butterknife is washed and the right command after.}
\end{minipage}\hfill
\begin{minipage}[t]{0.35\linewidth}
\vspace{0pt}
\centering
{\footnotesize\bfseries (b) Accuracy over all games}\\[2pt]
\includegraphics[width=\linewidth]{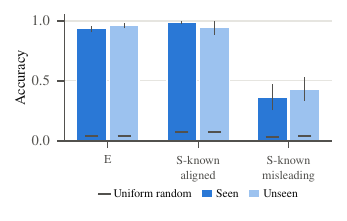}
\end{minipage}
\caption{\textbf{ALFWorld: \jev{} is right when the surface shortcut is right.} (a) \jev{}'s top choice at five decisions of one game, with probabilities averaged over two option orders. (b) Accuracy is how often \jev{} chooses an acceptable command. Error bars are 95\% bootstrap intervals over games, and dashes mark uniform random choice. The aligned/misleading split was defined on the unseen games and tested on the seen games.}
\label{fig:alfworld}
\end{figure}

\textbf{Results.} Figure~\ref{fig:alfworld}a follows one seen game. \jev{} gets the E steps right, such as taking and cleaning the butterknife. At the misleading step it goes to the drawer, which the task mentions, instead of the sinkbasin. After the butterknife is washed, the same command becomes correct, and \jev{} chooses it. Figure~\ref{fig:alfworld}b shows the same pattern over all games. On the seen games, \jev{} chooses an acceptable command at 93\% of E steps that take, put or clean (82\% of E steps), and at 99\% of aligned S-known steps. At misleading S-known steps it drops to 36\%, and it picks the lure 56\% of the time, five times the rate of uniform choice. The unseen games give similar results (Figure~\ref{fig:alfworld}b and Table~\ref{tab:alfworld-full} in Appendix~\ref{app:alfworld}). This confirms \textbf{predictions (i)--(iii)}. Why does \jev{} go straight to the drawer instead of first washing the butterknife at the sink?

\subsection{What Jev knows}
\label{sec:text-model}

Does \jev{} lack the order of subgoals, or fail to use it? ALFWorld writes each task in one of two forms, the \emph{ordered form} and the \emph{goal form} (Example~\ref{ex:forms}). Asked separately for the next subgoal, \jev{} answers correctly 100\% of the time under the ordered form but only 33\% under the goal form. The answer depends on the form. Under the ordered form, \jev{} knows the next subgoal, yet its decisions made in a single call are right only 68\% of the time. Writing its own answer into the state raises them to 86\%, as \textbf{prediction (v)} expects (Table~\ref{tab:probes}). Under the goal form, \jev{}'s own prediction is wrong 67\% of the time, and its wrong answers always skip to putting the object in place. Under the goal form, the order of subgoals must therefore come from code. When code adds the correct next subgoal to the state, accuracy rises to 92\% (Table~\ref{tab:probes}).

\begin{examplebox}{two ways to write one task}\label{ex:forms}
The \emph{ordered form} lists the steps, as in ``clean some butterknife and put it in drawer''. The \emph{goal form} only describes the final goal, as in ``put a clean butterknife in drawer''. The goal form states only one action, putting. Washing is implied only by the word ``clean''. Asked for the next subgoal, \jev{} answers ``clean'' under the ordered form, but under the goal form it usually answers ``put''.
\end{examplebox}

\begin{table}[t]
\centering
\small
\caption{\textbf{\jev{} can predict in a separate call and use a given prediction, but not both in one call.} $z$ is the missing prediction. Own $z$ is \jev{}'s answer from the separate call. Rows are the 90 misleading games and the 64 misleading ALFWorld steps of the seen games from task types that have both an ordered and a goal form.}
\label{tab:probes}
\input{tables/probes.tex}
\end{table}

\subsection{Code owns the simulation}
\label{sec:text-lookahead}

\textbf{Lookahead.} We now use code to perform the simulation. Before each decision, code runs every admissible command in a copy of the game. Each option then shows the command, what the game would print next and the commands that become possible only afterwards, such as ``then newly possible: clean butterknife 2 with sinkbasin 1'' (``Then:'' in Figure~\ref{fig:hero}c). This writes the missing prediction into each option. At misleading S-known steps, accuracy rises from 36\% to 79\% (Figure~\ref{fig:hero}a,b), and the rate of choosing the lure drops from 56\% to 21\%, as \textbf{prediction (iv)} expects.

\textbf{Whole games.} The same lookahead turns \jev{} into a strong controller for whole games (Table~\ref{tab:alfworld} in Appendix~\ref{app:alfworld}). \jev{} alone solves 42 of 134 unseen games. Showing what the game would print next raises this to 79, and adding the newly possible commands raises it to 116 (87\%). A word-overlap rule with the same lookahead information solves only 18. The gain therefore comes from \jev{}'s evaluation of the predicted outcomes, not from the lookahead information alone.

%% file: tables/probes.tex
\begin{tabular}{llcccc}
\toprule
 & & Predict & \multicolumn{3}{c}{Decide} \\
\cmidrule(lr){3-3}\cmidrule(lr){4-6}
Decisions & Missing $z$ & $z$ alone & no $z$ given & correct $z$ given & own $z$ given \\
\midrule
Games, misleading & Column's action & 0.90 & \textbf{0.33} & 0.98 & 0.94 \\
ALFWorld, ordered form & next subgoal & 1.00 & \textbf{0.68} & 0.86 & 0.86 \\
ALFWorld, goal form & next subgoal & 0.33 & \textbf{0.06} & 0.92 & 0.33 \\
\bottomrule
\end{tabular}

%% file: sections/7_dmc.tex
\section{Code Owns the Physics: Robot Control}
\label{sec:dmc}

Does the same division hold in robot control? Here the missing prediction is the physics. Whether a torque is good depends on how the arm will move next, and the state describes only where the arm is now. We therefore expect \jev{} to fail when it chooses raw torques. In contrast, we expect it to succeed when code performs the physics simulation and \jev{} only chooses among described outcomes. We test both cases on the same task, Stacker from the DeepMind Control Suite \citep{tassa2018deepmind}, where a planar arm with a two-finger gripper must bring a box to a target marker, a red square that shows where the box should go, either on the floor or one box above it.

\textbf{Without physics simulation.} We first use the native task, ``stack boxes so that one box sits at the target marker''. Every 0.02\,s, \jev{} sets the torques of the arm's five actuators, choosing for each among five described torque levels. \jev{}'s input describes the arm, gripper, boxes and joint angles. The input contains no outcome that code predicts by simulation, such as whether a move can be reached. In 20 episodes, \jev{} never picks up a box and solves none. Random torques also solve none (Table~\ref{tab:dmc}).

\textbf{With physics simulation.} We then give \jev{} a small library of manipulation skills, tuned on the native task (Appendix~\ref{app:dmc}). The skills move above a box, descend, push a crowded box aside, close or open the gripper, lift, place the held box, retreat and hold still. Code implements them with inverse kinematics \citep{tunyasuvunakool2020dm} and joint PD control \citep{lynch2017modern}. As a result, every skill corresponds to a physics simulation implemented by code. In other words, when \jev{} reads the skill options, it knows the outcome of that simulation. On the same 20 episodes of the native task, \jev{} now solves 9, as \textbf{prediction (iv)} expects, and the hand-written controller solves 8.

\begin{figure}[t]
\centering
{\footnotesize\textbf{Instruction:} ``Place box0 on top of box1, wherever box1 currently is, then move the hand away from the boxes. The target marker is irrelevant for this task.''}\\[4pt]
\setlength{\tabcolsep}{1pt}
\begin{tabular}{ccccc@{\hspace{5pt}}c@{\hspace{5pt}}c}
\multicolumn{5}{c}{\scriptsize\bfseries \jev{} choosing among skills that code simulates} & & \scriptsize\bfseries Hand-written \\[1pt]
\begin{tikzpicture}\node[anchor=south west,inner sep=0] (im) at (0,0) {\includegraphics[width=0.155\linewidth]{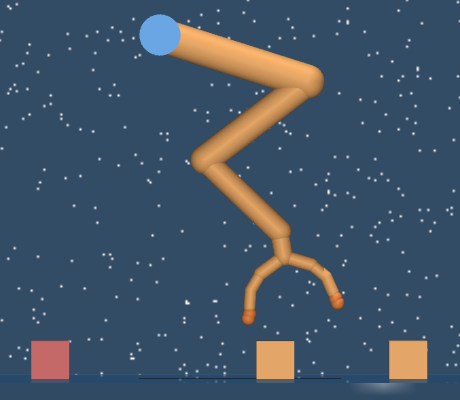}};\begin{scope}[x={(im.south east)},y={(im.north west)},font=\fontsize{5.5}{6}\selectfont,text=white,inner sep=0.4pt]\node at (0.11,0.26) {target};\node at (0.89,0.26) {box1};\node[anchor=east] (b0) at (0.45,0.125) {box0};\draw[white,->,line width=0.35pt] (b0.east) -- (0.545,0.125);\end{scope}\end{tikzpicture} & \includegraphics[width=0.155\linewidth]{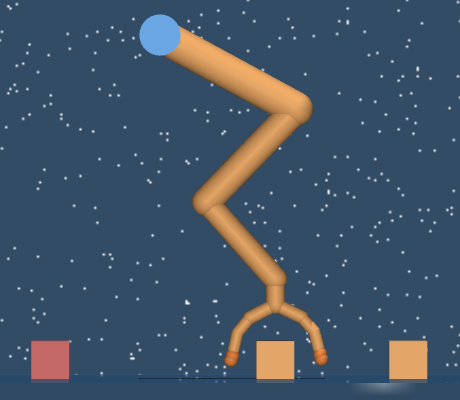} & \includegraphics[width=0.155\linewidth]{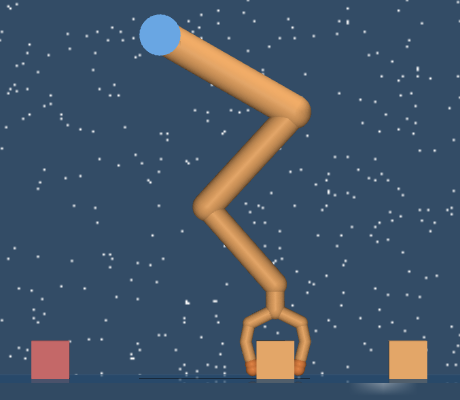} & \includegraphics[width=0.155\linewidth]{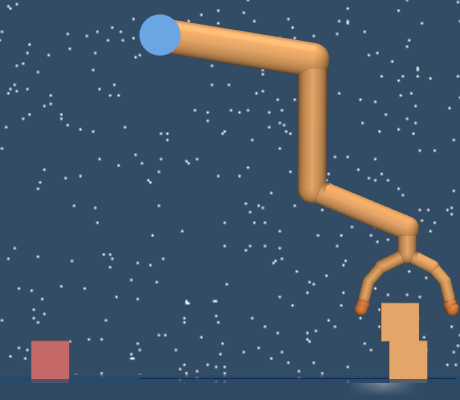} & \includegraphics[width=0.155\linewidth]{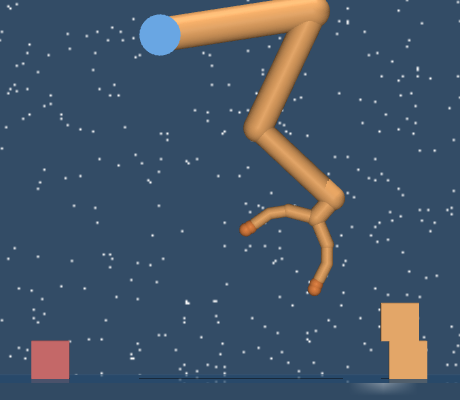} & \vrule width 0.4pt & \includegraphics[width=0.155\linewidth]{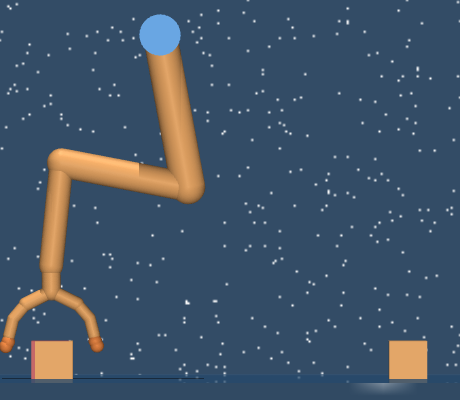} \\
\scriptsize\texttt{move\_above\_box0} & \scriptsize\texttt{descend\_to\_box0} & \scriptsize\texttt{close\_gripper} & \scriptsize\texttt{place\_on\_box1} & \scriptsize\texttt{retreat} & & \scriptsize\texttt{place\_at\_target} \\
\scriptsize Step 1: 0.96 & \scriptsize Step 2: 0.94 & \scriptsize Step 3: 0.79 & \scriptsize Step 4: 1.00 & \scriptsize Step 5: 0.71 & & \scriptsize Step 4 (\jev{}: 0.00) \\
\end{tabular}
\caption{\textbf{Stacker: \jev{} follows a new instruction by choosing among simulated skills.} Each frame shows the scene after one decision, with the chosen skill and \jev{}'s probability for it. Code executes and simulates every skill, and \jev{} never outputs a torque. On the same scene, the hand-written controller carries box0 to the red target marker at its fourth decision (right).}
\label{fig:dmc}
\end{figure}

\textbf{Generalization to new instructions.} To test generalization, we extend the native task with five new instructions, such as ``put box0 on box1'' or ``move the farther box to the target'', each run on 8 seeds with different starting conditions. We change the instruction text, while the skills are shared. \jev{} solves 22 of the 40 episodes, against 3 for the hand-written controller and none for random choice. We find that \jev{}'s evaluation ability, with skills developed for one task, provides promising generalization to unseen task variants. Figure~\ref{fig:dmc} shows one such episode. Although the settings with and without physics simulation differ in the action level, their contrast shows that code provides the simulation, so that \jev{} only needs to evaluate.

\begin{findingbox}{Contribution 3: remedy}
Code should own the simulation, and \jev{} the evaluation. With predictions or simulated skills supplied by code, \jev{} becomes an expert controller in games, ALFWorld and robot control.
\end{findingbox}

\begin{table}[t]
\centering
\small
\caption{\textbf{Stacker without and with physics simulation performed by code.} Success counts over episodes not already solved at reset. Random chooses uniformly among the same actions. The hand-written controller is the state machine for the native task, run unchanged.}
\label{tab:dmc}
\begin{tabular*}{\linewidth}{@{\extracolsep{\fill}}lccc}
\toprule
 & Raw torques, no simulation & \multicolumn{2}{c}{Skills simulated by code} \\
\cmidrule(lr){2-2}\cmidrule(lr){3-4}
Controller & Native task & Native task & 5 new instructions \\
\midrule
\jev{} & 0/20 (0\%) & \textbf{9/20 (45\%)} & \textbf{22/40 (55\%)} \\
Hand-written controller & -- & 8/20 (40\%) & 3/40 (8\%) \\
Random choice & 0/20 (0\%) & 1/20 (5\%) & 0/40 (0\%) \\
\bottomrule
\end{tabular*}
\end{table}

%% file: sections/2_related.tex
\section{Related Work}
\label{sec:related}

\textbf{Language models as policies over candidate actions.} A common way to use a language model as an agent is to let code list valid actions and let the model score them, as in SayCan \citep{ahn2022saycan} and admissible-action mapping \citep{huang2022language}. Text-game agents choose among TextWorld and ALFWorld commands \citep{cote2018textworld,shridhar2021alfworld}, with or without reasoning traces \citep{yao2023react,shinn2023reflexion}. GLAM \citep{carta2023grounding} and TWOSOME \citep{tan2024true} train such scorers with RL, and TALES collects text-adventure benchmarks \citep{cui2025tales}. \citet{li2026jevjudge} use \jev{} as a judge and escalate its low-confidence verdicts to a reasoning judge. They find it reliable when the verdict can be read from the text and weaker when it must be derived. These works report how often an agent succeeds. We instead ask \emph{which} decisions such a scorer gets right. The answer depends on whether a decision needs only evaluation or depends on simulation, which we determine from the environment, not from the model.

\textbf{Who simulates and who evaluates.} Tree of Thoughts, RAP and LATS let one model both propose future states and score them \citep{yao2023tree,hao2023reasoning,zhou2024language}. PlanBench shows that language models plan poorly on their own \citep{valmeekam2023planbench}, and LLM-Modulo pairs a generating model with external verifiers \citep{kambhampati2024position}. We use the opposite split. Code predicts the consequences of each option, and a judgment model without reasoning text evaluates them. In the terms of \citet{daw2005uncertainty}, \jev{} alone resembles model-free control, which acts on values learned from past experience instead of predicting consequences. Code adds the model-based part by predicting the consequences for it.

\textbf{Cognitive and strategic probes.} The CRT is built so that the intuitive answer is wrong \citep{frederick2005cognitive}. \citet{hagendorff2023human} found that GPT-3 models fall for it while ChatGPT-4 does not, and we borrow their error scoring and direct knowledge probes. Concurrent work pairs factual probes with action selection in agent loops and finds that most wrong actions of fast models follow a correct reading of the state \citep{li2026reading}. Other work applies cognitive tasks to language models \citep{binz2023using,codaforno2024cogbench}, and HANS splits items by whether a known heuristic answers them \citep{mccoy2019right}, which our aligned/misleading split mirrors. Behavioral game theory describes players by their depth of reasoning about others \citep{nagel1995unraveling,stahl1995players,camerer2004cognitive,costagomes2001cognition}, and human play departs from equilibrium in systematic ways \citep{goeree2001ten,arad2012money}. We vary depth and surface alignment separately and add a condition in which code states the opponent's action.

%% file: sections/8_discussion.tex
\section{Discussion}
\label{sec:discussion}

\textbf{The boundary lies inside a single call.} Across the four settings, \jev{} succeeds at evaluation when the input contains sufficient information. It fails when the right option depends on simulation to provide information missing from the input. When it fails, it picks the lure. In misleading matrix games, 90\% of its errors go to the action with the highest average payoff, against 50\% expected under random errors. At misleading ALFWorld steps, 89\% go to a command that mentions the task, against 12\% expected (Figure~\ref{fig:outcomes} in the appendix). In robot control, it never picks up a box when it must choose raw torques. Table~\ref{tab:probes} shows which part fails. Asked in a separate call, \jev{} predicts Column's action correctly in 90\% of misleading games. When code writes a correct prediction into the input, \jev{} chooses correctly in 98\% of them. It fails when one call must both perform the simulation and evaluate based on it, where it is right in only 33\%. This also explains why an instruction to reason does not help, since the instruction adds no prediction. Two limits remain. First, some facts are relations between options, such as one action paying more than another whatever the opponent does. Such a fact cannot be read from any single option. \jev{} mostly misses it, even when asked in a separate call (it identifies Column's dominated action in only 19\% of cases, Appendix~\ref{app:probes}), so code must compute it. Second, \jev{} follows the order in which the task mentions the actions, even when a word such as ``after'' says otherwise (Example~\ref{ex:after}).

\begin{examplebox}{a stated order that \jev{} does not follow}\label{ex:after}
``Put some butterknife in drawer after cleaning it'' mentions putting first, although cleaning must come first. With this wording, \jev{} decides correctly in 24\% of cases, against 69\% when the task mentions cleaning first (Appendix~\ref{app:probes}).
\end{examplebox}

\textbf{When can code make the prediction or simulation?} In RL terms, code supplies the world model that \jev{} lacks. With a simulator, such as the game rules, a copy of the game or a physics engine, code can compute the prediction directly. \jev{} then only evaluates. Without a simulator, code can still separate simulation and evaluation. It asks \jev{} for the prediction in one call and writes the answer into the next. This works when \jev{} can make the prediction from the input. When \jev{} lacks the ability to simulate a given environment, as with the unwashed knife, the prediction must come from elsewhere, for example a learned world model \citep{ha2018recurrent,hafner2020dream} or a larger reasoning model.

\textbf{Limitations.} We study one model and one API version. We have not tested other models that answer without reasoning text, nor how the boundary moves when reasoning is allowed. We therefore cannot yet say whether the effect is specific to \jev{}. The ALFWorld lookahead and the Stacker simulation use an exact copy of the environment, which a real agent may not have. Our ALFWorld lookahead looks one command ahead and lists the commands it makes possible. When a consequence appears only after several commands, code must search deeper, and the cost grows quickly with the number of commands. Our analysis of which predictions \jev{} can make from the input is partly post hoc.

\section{Conclusion}
\label{sec:conclusion}

We studied which decisions a judgment model can be trusted with. \jev{} succeeds when the right option needs only evaluation of the information in the input, and it fails when the right option depends on simulation, even when it can make that prediction in a separate call. The failure lies in a single call that must both simulate and evaluate. Letting code make the prediction or simulation closes this gap. Code can state the opponent's action, write back \jev{}'s own prediction, show the outcome of each command, or simulate each skill of a robot arm. With these simulations, \jev{} becomes an expert controller through its general evaluation ability, and in robot control it follows new instructions that a hand-written controller cannot. Whether the same boundary holds for other judgment models, and how far code can supply the simulation without an exact copy of the environment, are open questions. Code should own the simulation, and \jev{} the evaluation.

%% file: sections/appendix.tex
\section{The Jev Interface and Prompts}
\label{app:interface}

Every call is one request to the TypeSafe System One API with the model \texttt{jev-1.13.0} \citep{typesafe2026jev}. A request carries a JSON \texttt{state} and one choice question. The raw-torque condition of Stacker sends one question per actuator instead. A question consists of an instruction and a set of options, each with a key and a text description. In the CRT and ALFWorld the keys are neutral, such as \texttt{o0} and \texttt{c0}. In games they are the action labels A, B, C or X, Y, Z. In Stacker they are the names of the skills or torque levels, such as \texttt{place\_at\_target}. The response gives the chosen key, a probability for each option and a confidence score.

Each box below lists the state, the options and the instruction for one domain. A row named after a condition applies only to that condition.

\begin{promptspec}{CRT}
State & \texttt{question}: the question text. \\[2pt]
Options & Four candidate answers with keys \texttt{o0} to \texttt{o3}, for example ``\$0.05''. \\[2pt]
Instruction & Answer the question in \texttt{question}.
\end{promptspec}

\begin{promptspec}{Games}
State & \texttt{game}: ``one-shot simultaneous-move game, played once''. \texttt{you}: ``Row player: choose A, B or C''. \texttt{opponent}: ``Column player: chooses X, Y or Z at the same time as you, without seeing your choice. Column is rational, knows all payoffs, wants to maximize their own payoff, and knows that you are rational too.'' \texttt{payoffs\_row\_column}: the payoffs of both players in each cell, for example \texttt{"A,X": [82, 94]}. \\[2pt]
Options & Row's actions with keys A, B and C, for example ``Play A: you get 82 if Column plays X, 36 if Y, 20 if Z''. \\[2pt]
Instruction & You are the Row player. Choose the action that maximizes your own payoff against this opponent. \\[2pt]
Reason & Appends to the instruction: First work out which actions Column would never play given Column's own payoffs, then which of your actions is best against what remains, and repeat until one action is left. \\[2pt]
Told & Adds to the state: \texttt{opponent\_prediction}: ``Column will play X.''
\end{promptspec}

\begin{promptspec}{ALFWorld}
State & \texttt{task}: the task text. \texttt{current\_observation}: the game's last output. \texttt{history}: the earlier commands, each with the output it produced. \texttt{step}: the step number. \\[2pt]
Options & The admissible commands with keys \texttt{c0}, \texttt{c1}, \dots, for example ``go to sinkbasin 1''. \\[2pt]
Instruction & You control an agent in a household text game. Choose the next command that makes the most progress toward completing \texttt{task}. Use \texttt{history} to remember what you have already seen and done, avoid revisiting places you already checked, and pick up / use objects as the task requires. \\[2pt]
Lookahead v1 & Each option becomes ``command \texttt{->} output''. Appends to the instruction: Each option shows a command and, after `\texttt{->}', what the game would print right after it (computed by running the command in a copy of the game), marked when you have already seen exactly that output. \\[2pt]
Lookahead v2 & Each option also ends with ``then newly possible:'' and the new commands. Appends to the instruction: Each option shows a command and, after `\texttt{->}', what the game would print right after it and which commands would newly become possible (computed by running the command in a copy of the game), marked when you have already seen exactly that output.
\end{promptspec}
In v1 and v2, the mark is the tag ``[already seen]'' at the end of the option. It appears when the predicted output is identical to an output printed earlier in the same game.

\begin{promptspec}{Stacker with skills}
State & \texttt{task}: the task text. Target: \texttt{target\_xz\_m}, \texttt{target\_height}, \texttt{box\_under\_target}. Hand: \texttt{hand\_xz\_m}, \texttt{hand\_to\_target\_m}, \texttt{gripper}, \texttt{held\_box}, \texttt{held\_box\_at\_target}. Progress: \texttt{task\_complete}, \texttt{valid\_primitives\_now}, \texttt{last\_primitive}, \texttt{last\_outcome}. \texttt{boxes}: for each box \texttt{xz\_m}, \texttt{distance\_to\_target\_m}, \texttt{at\_target}, \texttt{under\_target}, \texttt{upright}, \texttt{gap\_to\_nearest\_box\_m}, \texttt{crowded}, \texttt{graspable\_from\_above}, \texttt{resting\_on\_arm}, \texttt{hand\_above\_it}, \texttt{between\_fingers}, \texttt{held}. Code computes all of these from the simulator. \\[2pt]
Options & The skills valid in the current gripper state, keyed by name: \texttt{move\_above\_}\emph{box}, \texttt{descend\_to\_}\emph{box}, \texttt{nudge\_}\emph{box}, \texttt{close\_gripper}, \texttt{open\_gripper}, \texttt{lift}, \texttt{place\_at\_target}, \texttt{place\_under\_target}, \texttt{retreat} and \texttt{hold\_still}. Each has a \texttt{what} field, which says what the skill does, and a \texttt{not\_for} field, which says when not to use it. For example, \texttt{place\_at\_target} has \texttt{what} ``Carry the held box to the target position and release it there'' and \texttt{not\_for} ``when no box is held, or the target is elevated and nothing supports it yet''. \\[2pt]
Instruction & Choose the next primitive for the robot arm so the stacking task is completed and rewarded as early as possible. Use \texttt{target\_height}, \texttt{box\_under\_target}, \texttt{boxes}, \texttt{held\_box} and \texttt{gripper}. \\[2pt]
New tasks & \texttt{task} holds the new instruction (Appendix~\ref{app:dmc}). Adds to the state: \texttt{box\_start\_xz\_m}, \texttt{box\_has\_visited\_target} and, while a box is held, \texttt{held\_box\_relations} (\texttt{at\_target}, \texttt{at\_start\_position}, \texttt{above\_}\emph{box}). Adds the skills \texttt{place\_at\_start} and \texttt{place\_on\_}\emph{box}. Replaces the instruction: Choose the next primitive for the robot arm so that the instruction in \texttt{task} is completed as early as possible. Use \texttt{boxes}, \texttt{held\_box}, \texttt{gripper}, \texttt{box\_start\_xz\_m} and \texttt{box\_has\_visited\_target}.
\end{promptspec}

\begin{promptspec}{Stacker with raw torques}
State & \texttt{rules}: A planar arm in the x-z plane with a root, shoulder, elbow and wrist joint and a two-finger gripper. Commands are normalized torques, not position targets. Gravity pulls the arm down, so a zero command lets it sag, and holding a pose needs a steady torque. Half torque held for 0.1\,s moves the hand by 2 to 8\,cm. A box is picked up by placing it between the open fingers and closing them. Boxes are 4.4\,cm cubes. The other fields are those of Stacker with skills, without \texttt{valid\_primitives\_now}, \texttt{last\_primitive} and \texttt{last\_outcome}, plus \texttt{joints}: the angle and velocity of each joint and the gripper opening. \\[2pt]
Options & One question for each actuator (root, shoulder, elbow, wrist, grasp). Five torque levels: \texttt{strong\_positive}, \texttt{gentle\_positive}, \texttt{none}, \texttt{gentle\_negative}, \texttt{strong\_negative}. For example, \texttt{strong\_negative} for the elbow reads ``Maximum torque that lowers the elbow angle, so the hand moves right (+x) and slightly up''. \\[2pt]
Instruction & Choose the torque command for the \emph{actuator} actuator for the next 0.02\,s so that the arm completes the instruction in \texttt{task} as early as possible. Use \texttt{joints}, \texttt{hand\_xz\_m}, \texttt{gripper} and \texttt{boxes}.
\end{promptspec}

Repeating an identical request changes the returned probabilities by 0.04 on average (maximum 0.12). It changes the chosen option in 12\% of cases. Averaging over option orders also reduces this noise.

\section{CRT Questions}
\label{app:crt}

We use the 150 questions of \citet{hagendorff2023human} (OSF \texttt{w5vhp}), 50 of each type. Type 1 follows the bat and ball, as in ``A pear and a fridge together cost \$140. The pear costs \$100 more than the fridge. How much does the fridge cost?''. Type 2 follows the machines that make widgets, and type 3 the doubling lily pads. Each question has a control version from the same materials, in which the intuitive answer is correct. In type 1 the control states the price of one item. In type 2 the number of workers stays the same. In type 3 the doubling clause is removed.

Each question has four options: the correct answer, the intuitive answer and two other wrong answers. In type 1 the other two are half the correct price and the price of the other item. In type 2 they are two of the product, difference or sum of the correct and intuitive times, or twice or three times the correct time. In type 3, where $t$ is the total time given in the story, the correct answer is $t-1$, the intuitive answer is $t/2$, and the other two are two of $t-2$, $t/4$, $t+1$ and $2t$.

Our first run reached 88\% on the control versions. The errors came from two ambiguities. First, a type-2 control asks ``How long does it take 7 people to tailor 4 jackets, if it takes 7 people 7 hours to tailor 7 jackets?''. The intended answer is 4 hours, but 7 hours is also right if each person makes a whole jacket. Second, the sentence we appended to type-3 controls to state a constant rate did not fit every story. A falling branch, for example, does not ``grow''. The reported run fixes both. It appends ``They share the work evenly, so any job can be split among all of them.'' to both versions of type-2 questions. This sentence does not change the answer of the lure version. It appends ``Assume it progresses at a constant rate, so equal amounts are added in equal periods of time.'' to type-3 controls. Controls then rose to 96\%, with type 2 going from 78\% to 92\% and type 3 from 85\% to 96\%. Most remaining control errors are type-2 questions such as asking how long 8 painters take to paint 1 painting when 8 painters paint 8 paintings in 8 hours. The intended answer is 1 hour, but splitting one painting among eight painters conflicts with common sense. The revision does not affect the main result. In both runs, \jev{} answers 99\% of the lure questions correctly.

\section{Games: Generation and Behavioral Models}
\label{app:pilot}

\begin{table}[h]
\centering
\small
\caption{\textbf{\jev{} plays as if the opponent acted at random (the logit level-1 rule).} One-parameter models $P(a) \propto \exp(\lambda\, v(a))$ are fitted to \jev{}'s probabilities over 630 calls per condition. The columns give the KL divergence from \jev{}'s distribution (lower is better), the fraction of \jev{}'s choices the model predicts, and the fitted $\lambda$ per 100 payoff units. Bold marks the best fit per condition.}
\label{tab:games-model}
\resizebox{\linewidth}{!}{\input{tables/games_model.tex}}
\end{table}

\textbf{Generation.} Games are $3 \times 3$. Each player's nine payoffs are distinct numbers drawn at random from 0--99, so no ties occur. Iterated elimination proceeds in rounds. In each round, and for both players simultaneously, it removes every action that is strictly dominated by another remaining pure action. Games are kept if a single outcome survives. The depth is the round, counted from zero, after which Row has one action left. Games are sampled until there are 30 of each of seven game types, namely depth 0 and depths 1 to 3 each aligned or misleading.

\textbf{Effect of depth.} We fit a logistic regression to the plain calls with $D \geq 1$:
\[
\operatorname{logit} P(\text{correct}) = \beta_0 + \beta_1 m + \beta_2 a + \beta_3\, m a + \beta_4 \mathbf{1}[D{=}2] + \beta_5 \mathbf{1}[D{=}3],
\]
where $m$ is the margin and $a$ is 1 in aligned games and 0 in misleading ones. The margin is the gap between the highest and the second-highest average payoff among Row's actions, where each average is taken over Column's three actions. Because $m$ and $a$ are in the model, $\beta_4$ and $\beta_5$ measure the effect of depth at the same margin and alignment. The margin lowers accuracy in misleading games ($\beta_1 = -9.8$ per 100 payoff units) and raises it in aligned ones ($\beta_1 + \beta_3 = 17.4$), as the surface shortcut predicts. Depth has no detectable effect ($\beta_4 = 0.22$ [$-0.52$, 1.03], $\beta_5 = -0.38$ [$-1.22$, 0.45]). Brackets give 95\% bootstrap intervals over games.

\textbf{Behavioral models.} We fit one-parameter softmax models $P(a) \propto \exp(\lambda\, v(a))$ to \jev{}'s probabilities, one for each decision rule $v$ (Table~\ref{tab:games-model}). In \emph{plain}, the level-1 rule fits best. It values each action by its average payoff against a random opponent. It has a KL divergence of 0.068 and predicts 77\% of \jev{}'s choices. Adding an indicator for the equilibrium to this rule barely improves the fit. Its fitted weight equals that of only 6 payoff units of average payoff, while the two best actions typically differ by more than 10 points. It also fits best under the \emph{reason} instruction (KL 0.049), with a lower $\lambda$ (3.9 against 5.5). In \emph{told}, the equilibrium rule fits best (KL 0.076) and predicts 99\% of choices.

\section{ALFWorld Details}
\label{app:alfworld}

\textbf{Expert and filtering.} We run ALFWorld's hand-coded expert, which searches receptacles in the order of ALFWorld's table of where each object type is usually found. We record every step until the game ends or 60 steps pass. The expert solves 122 of 134 unseen and 116 of 140 seen games. We keep only solved games, because the labels assume sensible expert steps.

\textbf{State tracking.} The labeler reads only the task and the history. From them it tracks the current location, the held object, which receptacles have been looked into, where each object was last seen, which objects have been cleaned, heated or cooled, and which have been placed. The task type fixes the subgoals. For example, in a clean task the agent must find the object, bring it to the sink and then bring it to the destination.

\textbf{Step types.} The type depends on the expert's command and the tracked state. A step is E if the command changes a goal fact: taking the object, cleaning, heating or cooling it, putting it in the destination, or using the desk lamp while holding the object. A step is S-known if the command changes no goal fact but goes to or opens a receptacle that a later goal step needs, such as the sink while holding an unwashed object, or returns to where the object was seen. A step is S-search if the needed object has not been seen yet and the command goes to or opens a receptacle to look for it. All other steps, such as looking around, closing a receptacle or discarding a wrong object, are labeled other and excluded.

\textbf{Surface shortcut.} A step is aligned if an acceptable command names the task's object or destination. It is misleading if no acceptable command does but another command does. Such commands are lures. Otherwise the step has no surface shortcut. We fixed this rule on the unseen games before testing it on the seen games.

\textbf{Acceptable sets.} For E steps, a command is acceptable if it has the same verb and object class as the expert's command. For a put command, the destination class must also match. This rule marks reasonable commands as wrong at E steps that heat, cool or use a lamp, such as opening a closed microwave before heating. We therefore report only E steps that take, put or clean, which make up 82\% of E steps. For S-known steps, going to a receptacle of the needed class is acceptable. So is going to where the needed object was seen, or opening the needed receptacle. For S-search steps, going to a receptacle not yet checked is acceptable. So is opening the current closed receptacle.

\begin{table}[h]
\centering
\small
\caption{\textbf{ALFWorld decisions on both splits.} The decisions are replayed from the expert's successful trajectories. The table gives the fraction of acceptable choices, with 95\% bootstrap intervals over games, and the rate under uniform random choice. The surface shortcut was defined on the unseen split and tested on the seen split.}
\label{tab:alfworld-full}
\begin{tabular}{lcccc}
\toprule
 & \multicolumn{2}{c}{Unseen (shortcut defined)} & \multicolumn{2}{c}{Seen (shortcut tested)} \\
Decision type & \jev{} & random & \jev{} & random \\
\midrule
E (take, put or clean) & 0.96 [0.94, 0.98] & 0.04 & 0.93 [0.90, 0.96] & 0.04 \\
S-known, aligned & 0.95 [0.89, 1.00] & 0.08 & 0.99 [0.97, 1.00] & 0.07 \\
S-known, misleading & 0.43 [0.33, 0.53] & 0.04 & 0.36 [0.26, 0.48] & 0.04 \\
S-search, aligned & 0.79 [0.72, 0.85] & 0.66 & 0.79 [0.74, 0.85] & 0.74 \\
S-search, misleading & 0.34 [0.24, 0.43] & 0.55 & 0.50 [0.36, 0.64] & 0.55 \\
S-search, no surface shortcut & 0.40 [0.33, 0.49] & 0.51 & 0.59 [0.40, 0.90] & 0.48 \\
\bottomrule
\end{tabular}
\end{table}

On search steps with no surface shortcut, \jev{} is below random choice on the unseen split (0.40 versus 0.51) and above it on the seen split (0.59 versus 0.48). We draw no conclusion from them.

\begin{table}[h]
\centering
\small
\setlength{\tabcolsep}{4pt}
\caption{\textbf{ALFWorld games solved in closed loop.} 134 unseen games, at most 50 steps. The columns from Pick to Two give the fraction solved for each task type in the first run, and Run 2 repeats the \jev{} conditions. Stuck counts failed games whose last ten commands contain at most three distinct commands. Tokens are \jev{}'s input tokens. The game copy shows what each command would reveal, so these numbers are not comparable to agents without a simulator.}
\label{tab:alfworld}
\resizebox{\linewidth}{!}{\input{tables/alfworld_lookahead.tex}}
\end{table}

\textbf{Lookahead.} ALFWorld keeps its game state inside a C library, so the game cannot be copied from Python. At each decision, the process holding the game forks once per admissible command. Each child runs its command, sends back the game's output, the new admissible commands and the win flag, and exits, so the parent's game is left untouched. A random policy with a fixed seed produces identical trajectories with and without lookahead, so the lookahead does not change the game. The predicted outputs also match the real ones at every step. Per decision, the forked lookahead took about 3\,s and the \jev{} call about 0.4\,s. Table~\ref{tab:alfworld} gives the closed-loop results. Lookahead v1 solves 41 games that \jev{} alone fails and fails 4 that it solves (exact McNemar test, $p = 9 \times 10^{-9}$). Lookahead v2 gains 38 games over v1 and loses 1 ($p = 1.5 \times 10^{-10}$). The word-overlap rule uses no model. It scores each option, the command together with its predicted output, by the words it shares with the task and picks the best option whose output was not already seen.

\section{DMC Details}
\label{app:dmc}

\textbf{Stacker skills.} Each skill computes target joint angles and drives the arm there with a PD (proportional-derivative) controller at 100\,Hz, with gains $k_p = (21, 14, 7, 3.5)$ and $k_d = (2.1, 1.4, 0.7, 0.35)$. The target angles come from inverse kinematics (IK), which finds joint angles that put the hand at a given position. IK is run from several random starting points and can return several solutions. Code simulates each solution in a copy of the physics and keeps the one that moves the other box least and needs the least joint motion. The place skills run as one unit: carry the box, release it and wait for it to settle. An episode succeeds if the task's success check holds at its end. For the native task, the check requires a box at the target and the hand at least 10\,cm away. Once the check holds, code offers only \emph{hold still}, so a finished task is not undone. Episodes last 10\,s on the native task and 20\,s on the new instructions.

\textbf{Instructions.} The native task reads ``Stack boxes so that one box sits at the target marker. Reward accrues only while a box is within about one box width of the target AND the hand is at least 10 cm away from the target.'' For each new instruction, we redraw the initial scene, up to a fixed number of times, until the task can be done, for example until each needed box rests on the floor with room around it and can be reached from above. This is similar to how dm\_control redraws scenes in which objects overlap. We take seeds in order from 0 and skip scenes where the instruction already holds at the start or no workable scene is found, until 8 scenes remain. The five new instructions read as follows.
\begin{itemize}
\item ``Place box0 on top of box1, wherever box1 currently is, then move the hand away from the boxes. The target marker is irrelevant for this task.''
\item ``Place box1 on top of box0, wherever box0 currently is, then move the hand away from the boxes. The target marker is irrelevant for this task.''
\item ``Bring box0 to the target marker on the floor, then carry it back to where it started (see box\_start\_xz\_m), release it there and move the hand away.''
\item ``Bring box1 to the target marker (on the floor at the target x) and move the hand away. Do not move box0 at all.''
\item ``Take whichever box started farther from the target marker and bring it to the target (on the floor at the target x), then move the hand away. Leave the other box alone.''
\end{itemize}

\textbf{Stacker without simulation.} Every 0.02\,s, \jev{} answers one question per actuator, choosing among five torque levels. Each level is described by the direction in which it moves the hand. We measured these directions once, by applying half torque for 0.1\,s from the arm's start pose. We run seeds 0 to 21. In seeds 7 and 9 a box already sits at the target when the episode starts, so we exclude them, which leaves 20 episodes. In these, over about 440 decisions each, \jev{} never holds a box. On average, the closest its hand comes to a box is 25\,cm, against 16\,cm with random torques. The skill runs in Table~\ref{tab:dmc} use the same seeds.

\section{Knowledge Probes}
\label{app:probes}

\textbf{Game questions.} The probes use the 180 games with $D \geq 1$, each in three orders of Row's actions, with the state of the plain condition. The prediction question asks ``Which action will Column play?''. Its options describe Column's payoffs for each of its actions. An elimination question asks which of one player's remaining actions is strictly dominated, that is, pays less than another remaining action of the same player against every remaining action of the other player. Its options are the remaining actions and ``none''. In the elimination write-back, code asks Column and Row in turn and writes each answer into the state as a note, such as ``Column will never play Z.''. It stops when Row has one action left or both players answer ``none''. The prediction write-back writes \jev{}'s own prediction into the state as ``Column will play Y.'', exactly as in the told condition.

\textbf{Game results.} When Column had a dominated action, \jev{} named it in 19\% of steps and wrongly answered ``none'' in 81\%. When Column had none, it answered correctly in 98\%. For Row, the same two accuracies were 58\% and 85\%. Overall, single elimination steps are correct in 57\% of calls. With correct elimination notes, decisions in misleading games reach 94\%, but with \jev{}'s own elimination steps written back they reach only 43\%. In the prediction probe, \jev{} picks the column with the highest average payoff to Column in 88\% of calls. In the 17 misleading games where that column is not the equilibrium, it names Column's action correctly only 55\% of the time. The 244 calls with a correct prediction and the 26 with a wrong one give the decision accuracies of 32\% and 35\% in Section~\ref{sec:games-probe}.

\textbf{ALFWorld questions.} The subgoal question reads ``Which of these must happen next to complete \texttt{task}? Use \texttt{history} to see what has already been done.''. Its six options are to find and pick up the object, clean it, heat it, cool it, put it in the destination, and turn on the desk lamp while holding it. The labeler of Appendix~\ref{app:alfworld} gives the correct answer. The told condition writes the correct subgoal into the state as a field \texttt{next\_subgoal}, and the write-back writes \jev{}'s own answer there. Both then ask the decision question of Section~\ref{sec:text-diag} unchanged.

\textbf{ALFWorld results.} At the 149 aligned S-known steps, where the next subgoal is to put the object, predicting the subgoal, deciding with the correct subgoal given and deciding with \jev{}'s own subgoal written back all reach 100\%. At misleading steps, \jev{} names the subgoal correctly in 63\% of calls, and every error names putting the object. By task type, it is right in 77\% of clean, 57\% of heat and 50\% of cool steps. With the correct subgoal given, clean and heat steps reach 100\% and cool steps 70\%. On cool steps \jev{} then often goes to the sink or to the destination. Table~\ref{tab:probes} uses the 64 misleading seen-split steps at which the object must still be cleaned, heated or cooled, 28 in the ordered form and 36 in the goal form, each in two option orders. The other 6 misleading steps belong to lamp and two-object tasks, which have neither form. The game row of Table~\ref{tab:probes} uses the 90 misleading games in three orders.

\begin{figure}[h]
\centering
\includegraphics[width=0.7\linewidth]{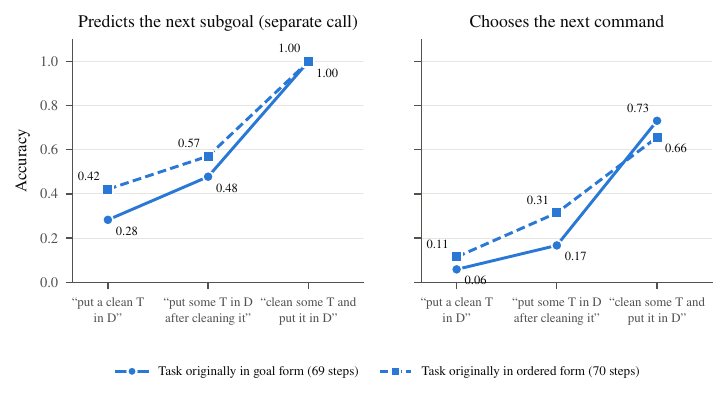}
\caption{\textbf{\jev{} knows what comes first only when the task mentions it first.} Misleading ALFWorld steps of both splits under three task templates. Code swaps the template and changes nothing else. Left: \jev{} predicts the next subgoal in a separate call. Right: \jev{} chooses a command.}
\label{fig:wording}
\end{figure}

\textbf{Task wording.} ALFWorld writes each clean, heat or cool task in the ordered form or the goal form (Example~\ref{ex:forms}), using fixed templates such as ``heat some T and put it in D.'' and ``put a hot T in D.'', where T is the object and D the destination. Code swaps only the template. This moves decisions from 6\% to 73\% when the goal form becomes the ordered form, and from 66\% to 11\% the other way, and the subgoal answers flip with them (Figure~\ref{fig:wording}). A third template, ``put some T in D after cleaning it'', says with ``after'' that cleaning comes first but mentions putting first. With it, decisions reach only 24\% (Example~\ref{ex:after}).

\section{Error Outcomes}
\label{app:errors}

\begin{figure}[h]
\centering
\includegraphics[width=\linewidth]{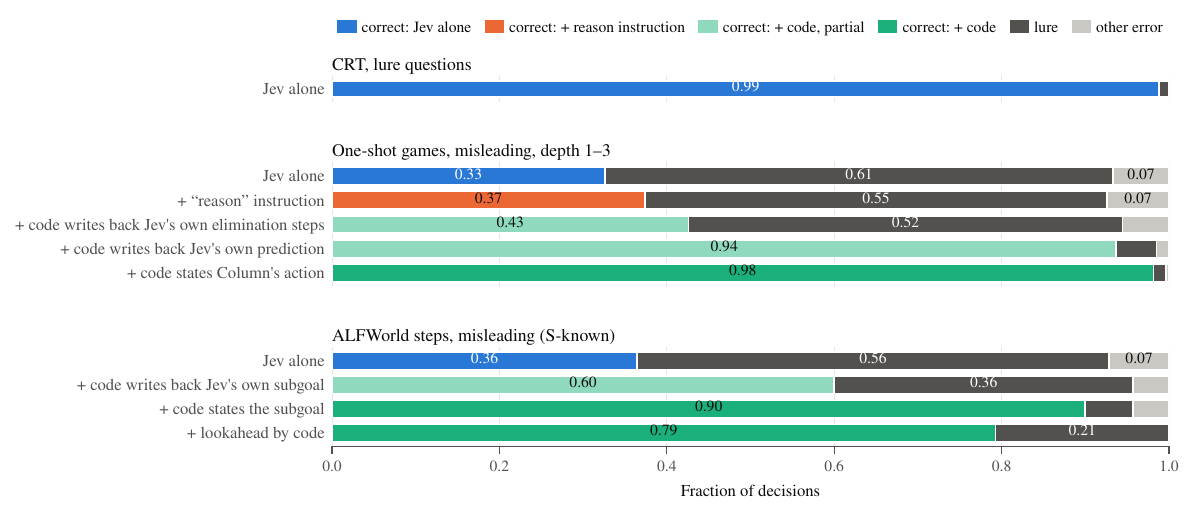}
\caption{\textbf{When \jev{} is wrong, it picks the lure.} The decisions with a lure in each domain, scored as correct, lure or other error \citep{hagendorff2023human}. The panels cover the 150 CRT questions, the 90 misleading games with $D \geq 1$ and the 70 misleading S-known decisions of the ALFWorld seen split. The lure is the intuitive answer in the CRT and the action with the highest average payoff in games. In ALFWorld it is any wrong command that mentions an object or place named in the task description. The color of the correct segment marks the condition, as in the other figures. Light aqua marks partial help from code. Here code writes back \jev{}'s own answer to a separate question. Rates are averaged over items and option orders.}
\label{fig:outcomes}
\end{figure}

Figure~\ref{fig:outcomes} scores every decision that has a lure as correct, as the lure, or as another error \citep{hagendorff2023human}. If \jev{}'s errors were spread evenly over the wrong options, the lure would receive one in three errors in the CRT, one in two in games and about one in eight in ALFWorld. Instead, almost all errors go to the lure. In the CRT, the few errors (1\%) all pick the intuitive answer. In misleading games, 90\% of errors pick the action with the highest average payoff. At misleading S-known ALFWorld steps, 89\% pick a command that mentions the task. When code supplies the missing prediction, most decisions move from the lure to the correct option. Stating Column's action leaves 2\% errors in games, and the lookahead cuts the lure rate in ALFWorld from 56\% to 21\%.

\section{Compute and Cost}
\label{app:compute}

All \jev{} calls use the public API, which charges only for input tokens. Table~\ref{tab:tokens} lists the input tokens of each experiment. In total, the reported runs used about 94M input tokens. At the price TypeSafe listed in September 2026, \$0.042 per million input tokens, they cost about \$4. The two closed-loop ALFWorld runs account for about \$2.2 of this, and the Stacker runs for about \$0.9. The environments ran on the CPUs of a single server.

\begin{table}[H]
\centering
\small
\caption{\textbf{Input tokens of each experiment.}}
\label{tab:tokens}
\begin{tabular}{lr}
\toprule
Experiment & Input tokens \\
\midrule
CRT & 0.73M \\
Games & 1.18M \\
Game probes & 4.1M \\
ALFWorld subgoal probe & 1.3M \\
ALFWorld template swaps & 1.1M + 0.7M \\
ALFWorld decisions: unseen / seen / seen with lookahead & 4.8M / 4.5M / 3.4M \\
ALFWorld whole games, two runs each: alone / v1 / v2 & $2 \times 10.6$M / $2 \times 9.2$M / $2 \times 5.9$M \\
Stacker with skills: native task / new instructions & 0.33M / 0.98M \\
Stacker with raw torques & 19.6M \\
\midrule
Total & $\approx$ 94M \\
\bottomrule
\end{tabular}
\end{table}

\section{Use of Large Language Models}
The authors conceived the research idea, designed the experiments and set the argument of the paper. They decided which experiments to run and report, and they shaped the logic and wording of every section through iterative review. An AI coding assistant, Claude Code (Anthropic), implemented most of the experiment and analysis code and ran the experiments under the authors' direction. It also drafted text, which the authors revised sentence by sentence. Another assistant, OpenAI's Codex, reviewed drafts for logic and consistency. Every reference was checked against its published source. The authors checked every number against the raw runs and take full responsibility for the content. \jev{}, the model we study, is part of the method and is described in the main text, not here.

%% file: tables/games_model.tex
\begin{tabular}{lcccccc}
\toprule
 & \multicolumn{3}{c}{\emph{plain}} & \multicolumn{3}{c}{\emph{told}} \\
Decision rule & KL & choice & $\lambda$ & KL & choice & $\lambda$ \\
\midrule
Level-1: average own payoff & \textbf{0.068} & 0.77 & 5.5 & 0.771 & 0.57 & 6.2 \\
Maxmax: best own payoff & 0.137 & 0.66 & 4.2 & 0.919 & 0.50 & 3.4 \\
Maxmin: worst own payoff & 0.172 & 0.66 & 3.0 & 0.889 & 0.55 & 3.3 \\
Level-2: best reply to a level-1 opponent & 0.192 & 0.65 & 1.9 & 0.485 & 0.89 & 7.5 \\
Equilibrium action & 0.203 & 0.65 & 0.8 & \textbf{0.076} & 0.99 & 4.2 \\
Payoff given the stated action & 0.194 & 0.65 & 2.0 & 0.103 & 0.99 & 26.9 \\
Uniform & 0.286 & -- & -- & 1.028 & -- & -- \\
\bottomrule
\end{tabular}

%% file: tables/alfworld_lookahead.tex
\begin{tabular}{lccccccccccc}
\toprule
Condition & Solved & 95\% CI & Run 2 & Pick & Look & Clean & Heat & Cool & Two & Stuck & Tokens \\
\midrule
Jev alone & 42/134 & [0.24, 0.40] & 39/134 & 0.58 & 0.56 & 0.19 & 0.09 & 0.05 & 0.53 & 54/92 & 10.6M \\
Jev + lookahead v1 (next observation) & 79/134 & [0.51, 0.67] & 84/134 & 0.96 & 0.72 & 0.42 & 0.39 & 0.19 & 1.00 & 45/55 & 9.2M \\
Jev + lookahead v2 (+ new commands) & \textbf{116/134} & [0.81, 0.92] & 117/134 & 0.96 & 0.94 & 0.77 & 0.78 & 0.81 & 1.00 & 14/18 & 5.9M \\
Word-overlap rule + lookahead v2 & 18/134 & [0.08, 0.19] & -- & 0.17 & 0.44 & 0.13 & 0.04 & 0.05 & 0.00 & 1/116 & -- \\
Word-overlap rule + lookahead v1 & 7/134 & [0.01, 0.09] & -- & 0.12 & 0.22 & 0.00 & 0.00 & 0.00 & 0.00 & 1/127 & -- \\
Random & 1/134 & [0.00, 0.02] & -- & 0.00 & 0.06 & 0.00 & 0.00 & 0.00 & 0.00 & 0/133 & -- \\
\bottomrule
\end{tabular}